\documentclass{article}

    \usepackage[preprint,nonatbib]{neurips_2019}

\usepackage[utf8]{inputenc} 
\usepackage[T1]{fontenc}    
\usepackage{hyperref}       
\usepackage{url}            
\usepackage{booktabs}       
\usepackage{amsfonts}       
\usepackage{nicefrac}       
\usepackage{microtype}      

\usepackage{times}
\usepackage{latexsym}
\usepackage{tabu,multirow}

\usepackage{multirow}
\usepackage{latexsym}
\usepackage{mathrsfs}
\usepackage{amsfonts}
\usepackage{amssymb}
\usepackage{amsmath}
\usepackage{bm}
\usepackage{url}
\usepackage{times}
\usepackage{mydef}
\usepackage{multirow, tabularx}
\usepackage{natbib}

\usepackage{graphicx} 
\graphicspath{ {.} }

\usepackage{booktabs}
\usepackage{amsmath}
\usepackage{color}
\usepackage{algorithm,algorithmicx,algpseudocode}
\usepackage{tabu}
\usepackage{pgfplots}

\title{DropAttention: A Regularization Method for
Fully-Connected Self-Attention Networks}

\author{
  \AND
  Lin Zehui\\
  Fudan University\\
  \texttt{linzh18@fudan.edu.cn}\\
  \And
  Pengfei Liu \thanks{Co-mentoring}\\
  Fudan University\\
  \texttt{pfliu14@fudan.edu.cn}\\
  \And
  Luyao Huang\\
  Fudan University\\
  \texttt{lyhuang18@fudan.edu.cn}\\
  \And
  Junkun Chen\\
  Fudan University\\
  \texttt{jkchen16@fudan.edu.cn}\\
  \And
  Xipeng Qiu \thanks{Corresponding author}\\
  Fudan University\\
  \texttt{xpqiu@fudan.edu.cn}\\
  \And
  Xuanjing Huang \\
  Fudan University \\
  \texttt{xjhuang@fudan.edu.cn}\\
}

\begin{document}

\maketitle

\begin{abstract}
Variants dropout methods have been designed for the fully-connected layer, convolutional layer and recurrent layer in neural networks, and shown to be effective to avoid overfitting. As an appealing alternative to recurrent and convolutional layers, the fully-connected self-attention layer surprisingly lacks a specific dropout method.
This paper explores the possibility of \textit{regularizing the attention weights} in Transformers to prevent different contextualized feature vectors from co-adaption.
Experiments on a wide range of tasks show that DropAttention can improve performance and reduce overfitting.

\end{abstract}

\section{Introduction}

As an effective and easy-to-implement regularization method, Dropout has been first designed for fully-connected layers in neural models \cite{srivastava2014dropout}.
Over the past few years, a host of variants of dropout have been introduced.
For recurrent neural networks (RNNs), dropout is only applied to the input layers before the successful attempt in \cite{krueger2016zoneout,Semeniuta2016,gal2016theoretically}. Also, a dozen of dropout methods for convolutional neural networks (CNNs) have been proposed in \cite{tompson2015efficient,huang2016deep,larsson2016fractalnet,gastaldi2017shake,ghiasi2018dropblock,
zoph2018learning,yamada2018shakedrop}.
On the other hand, fully-connected self-attention neural networks, such as Transformers \cite{vaswani2017attention}, have emerged as a very appealing alternative to RNNs and CNNs when dealing with sequence modelling tasks.

Although Transformers incorporate dropout operators in their architecture, the regularization effect of dropout in the self-attention has not been thoroughly analyzed in the literature.

The success of the adaption of dropout for fully-connected, convolutional and recurrent layers gives us a tantalizing hint that a more specific dropout for self-attentional operators might be needed.
Additionally, the original publicly code\footnote{https://github.com/tensorflow/tensor2tensor} of Transformer \cite{vaswani2017attention} with Dropout trick also provides the evidence for this work, although it's less understood why it works and how it might be extended. 
In this paper,  we demonstrate the benefit of dropout in self-attention layers (DropAttention) with two key distinctions compared with the dropout used in fully-connected layers and recurrent layers.
The first is that DropAttention randomly sets attention weights to zero, which can be interpreted as dropping a set of neurons along different dimensions.
Specifically, DropAttention aims to encourage the model to utilize the full context of the input sequences rather than relying solely on a small piece of features.
For example, for sentiment classification, the prediction is usually dominated by one or several emotional words, ignoring other informative patterns. This can make the model overfit some specific patterns.
In fully-connected and recurrent layers, dropout discourages the complex co-adaptation of different units in the same layer, while DropAttention prevents different contextualized feature vectors from co-adapting, learning features which are generally helpful for task-specific prediction.
Secondly, in addition to dropping out individual attentional units, we also explore the possibility of operating in contiguous regions.
It is inspired by DropBlock  \cite{ghiasi2018dropblock} where units in a contiguous region of a convolutional feature map are discarded together. It is a more effective way of dropping for attention layers, since a semantic unit are usually composed of several spatially consecutive words.
Experiments on a wide range of tasks with different-scale datasets show that DropAttention can improve performance and reduce overfitting.

\section{Related Work}

\begin{table*}[t!]
\center \footnotesize
\tabcolsep0.07in
\begin{tabular}{l*3{l}}
\toprule

\textbf{Methods}
&  \textbf{Dropped Objects}
&\textbf{Spaces} & \textbf{Layers}\\

\midrule
Dropout  \cite{srivastava2014dropout}             &  Unit    & Hidden    &  FCN   \\

DropConnect  \cite{wan2013regularization}    & Weight    & Hidden & FCN      \\
SpatialDropout  \cite{kalchbrenner2014convolutional}       &	 Unit     & Hidden & CNN   \\

Cutout  \cite{devries2017improved}            &  Unit     & Input    &  CNN   \\

DropEmb  \cite{gal2016theoretically}        &  Weight   & Input    &  Lookup   \\
Variational Dropout \citep{gal2016theoretically}  &  Unit    & Hidden    &  RNN   \\
Zoneout  \cite{krueger2016zoneout}          &  Unit    & Hidden    &  RNN   \\

DropBlock  \cite{ghiasi2018dropblock}       &  Region of Units     & Hidden    &  CNN   \\
\midrule
DropAttention   &  Region of Weights   & Input\& Hidden    &  Self-Attention   \\
\bottomrule
\end{tabular}
\caption{
A comparison of published methods for dropout. ``\texttt{Unit}'' denotes the neuron of a hidden vector while ``\texttt{Weight}'' represents the learnable parameter or attention score.  ``\texttt{FCN}'' refers to the fully-connected layer.
} \label{tab:all-models}
\end{table*}

We present a summary of existing models by highlighting differences among \textit{dropped object}, \textit{spaces} and \textit{layers} as shown in Table~\ref{tab:all-models}. The original idea of Dropout is proposed by \cite{srivastava2014dropout} for fully-connected networks, which is regarded as an effective regularization method. After that, many dropout techniques for specific network architectures, such as CNNs and RNNs, have been proposed. For CNNs, most successful methods require the noise to be structured \cite{tompson2015efficient,huang2016deep,larsson2016fractalnet,gastaldi2017shake,ghiasi2018dropblock,
zoph2018learning,yamada2018shakedrop}.
For example, SpatialDropout\cite{kalchbrenner2014convolutional}  is used to address the spatial correlation problem.
DropConnect \cite{wan2013regularization} sets a randomly selected subset of weights within the network to zero.
For RNNs, Variational Dropout \cite{gal2016theoretically} and ZoneOut \cite{krueger2016zoneout} are most widely used methods. In Variational Dropout, dropout rate is learned and the same neurons are dropped at every timestep. In ZoneOut, it stochastically forces some hidden units to maintain their previous values instead of dropping.
Different from these methods, in this paper, we explore how to drop information on self-attention layers.

\section{Background}


\subsection{Transformer Architecture}
The typical fully-connected self-attention architecture is the Transformer \cite{vaswani2017attention}, which uses the scaled dot-product attention to model the intra-interactions of a sequence. Given a sequence of vectors $H \in \mathbb{R}^{l\times d}$, where $l$ and $d$ represent the length of and the dimension of the input sequence, the self-attention projects $H$ into three different matrices: the query matrix $Q$, the key matrix $K$ and the value matrix vector $V$, and uses scaled dot-product attention to get the output representation.
\begin{align}
    &Q,K,V = H W^Q, H W^K,H W^V \label{eq:qkv}\\
    &\mathrm{Attn}(Q,K,V) = \mathrm{softmax}(\frac{Q K^T}{\sqrt{d_k}}) V,
\end{align}
where $ W ^Q,W^K,W^V \in \mathbb{R}^{d\times d_k}$ are learnable parameters and $\mathrm{softmax}()$ is performed row-wise.

To enhance the ability of self-attention, multi-head self-attention is introduced as an extension of the single head self-attention, which jointly model the multiple interactions from different representation
spaces,
\begin{align}
    &\mathrm{MultiHead}(H) =  [ \text{head}_1;...;\text{head}_{k} ] W^O,\\
    &\mathrm{where} \quad \text{head}_i = \mathrm{Attn}(H W^Q_i,H W^K_i,H W^V_i),
\end{align}
where $W^O,W^Q_i,W^K_i,W^V_i (i\in [1,h])$ are learnable parameters.
Transformer consists of several stacked multi-head self-attention layers and fully-connected layers.  Assuming the input of the self-attention layer is $H$, its output $\tilde{H}$ is calculated by
\begin{align}
    Z =& H + \mathrm{MultiHead}(\textrm{layer-norm}(H)), \\
    \tilde{H} =& Z + \mathrm{MLP}(\textrm{layer-norm}(Z)), \label{eq:residue}
\end{align}
where $\mathrm{layer-norm}(\cdot)$ represents the layer normalization \cite{ba2016layer} .

Besides, since the self-attention ignores the order information of a sequence, a positional embedding $PE$ is used to represent the positional information.

\subsection{Dropout}

Dropout \cite{srivastava2014dropout} is a popular regularization form for fully-connected neural network. It breaks up co-adaptation between units, therefore it can significantly reduce overfitting and improve test performance.
Each unit of a hidden layer $\bh^{(l)}\in \mathbb{R}^d$ is dropped with probability $p$ by setting it to 0.
\begin{align}
\bh^{(l+1)} = f(W \bm\odot \bh^{(l)}),
\end{align}
where $\bm \in \{0,1\}^d$ is a binary mask vector with each element $j$ drawn independently from $m_j \sim Bernoulli(p)$, and $\odot$ denotes element-wise production.

DropConnect \cite{wan2013regularization} is a generalization of Dropout. It randomly drops hidden layers weights instead of units. Assume $M$ is a binary mask matrix drawn from $M_{ij}\sim Bernoulli(p) $, $W$ is the hidden layer weights. Then DropConnect can be formulated as,
\begin{align}
\bh^{(l+1)} = f((W\odot M) \bh^{(l)})
\end{align}

Dropout essentially drops the entire column of the weight matrix. Therefore, Dropout can be regarded as a special case of DropConnect, where a whole column weight is dropped.

Since Dropout and DropConnect achieve great success on fully-connected layer, a natural motivation is whether a specific dropout method is needed for the fully-connected self-attention networks. Experiments conducted shows that a new dropout method designed for fully-connected self-attention networks can also reduce overfitting and obtain improvements.

\section{DropAttention}

\begin{figure}[t]

\centering
  \includegraphics[scale=0.25]{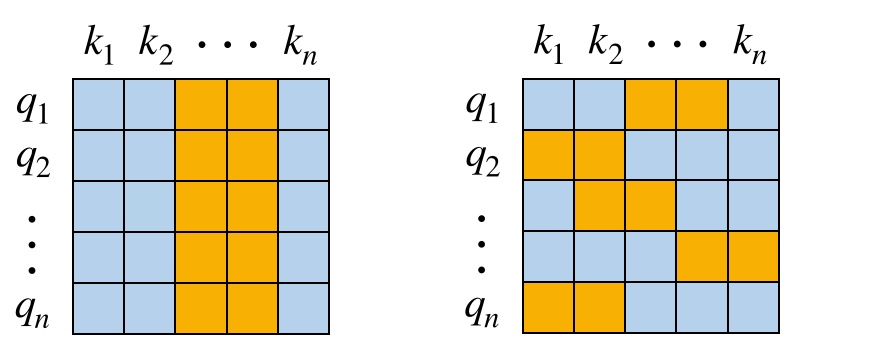}
  
  (a) DropAttention(c) \hspace{2.5em}  (b) DropAttention(e)
  \caption{Illustration of DropAttentions over a $5\times 5$ attention weight matrix. The ``yellow'' elements are dropped. The size of drop window is $w=2$ and drop rate is $p=0.4$.}
  \label{fig:dropattention}
\end{figure}

In this section, we will introduce our attention regularization method: DropAttention.

Given a sequence of vectors $H \in \mathbb{R}^{l\times d}$, the fully-connected self-attention layer can be reformulated into
\begin{align}
\tilde{H} = f(\Lambda V),
\end{align}
where $\Lambda= \mathrm{softmax}(\frac{Q K^T}{\sqrt{d_k}})$, $f(\cdot)$ is a residual nonlinear function defined by Eq. \eqref{eq:residue} and $Q,K,V$ is calculated by Eq. \eqref{eq:qkv}.

The output of $i$-th position is
\begin{align}
\tilde{\bh}_i = f(\sum_{j=1}^{l}\lambda_{ij} \bv_j),
\end{align}
where $\tilde{\bh}_i$ is the $i$-th row vector of $\tilde{H}$ and $\bv_j$ is the $j$-th row vector of $V$. $\lambda_{ij}$ is the entry of $\Lambda$.

With this formulation, we can connect the self-attention layer to the fully-connected layer with two differences. The first difference is the weight matrix $\Lambda$ is dynamically generated. The second difference is that the basic unit is a vector rather than a neuron.

Due to the similarity between fully-connected layer and self-attention layer, we can introduce the popular dropout methods for FCN to self-attention mechanism.
In detail, we propose two dropout methods for the fully-connected self-attention layer: DropAttention(c) and DropAttention(e).

1) \textbf{DropAttention(c)} means to randomly drop ``column'' in attention weight matrix, which is a simple method similar to the standard Dropout \cite{srivastava2014dropout}. We randomly drop the unit $\bv_j, 1\leq j\leq l$. Note that $\bv_j$ here is a vector instead of a single neuron.

2) \textbf{DropAttention(e)} means to randomly drop ``element'' in attention weight matrix, which is a more generalized method of the DropAttention(c). Similar to DropConnect  \cite{wan2013regularization}, DropAttention(e) randomly drops elements in attention weights matrix $\Lambda$.
DropAttention(c) can be regarded as a special case of DropAttention(e) in which a whole column of $\Lambda$ is dropped.

Besides the basic dropping strategies, we also augment the DropAttentions with two functions.

\subsection{Dropping Contiguous Region}
Inspired by DropBlock \cite{ghiasi2018dropblock}, we drop contiguous region of the attention weights matrix instead of independent random units. The behind motivation is based on distributional hypothesis \cite{harris1954distributional}: words that are used and occur in the same contexts tend to purport similar meanings. In Transformer \cite{vaswani2017attention} where multi-layer structure is used, when dropping independent random units, information correlated with the dropped input can still be restored in the next layer through surrounding words, which may cause the networks overfitting. Dropping the whole semantic unit consisting of several words can be a more effective way of dropping out.

Therefore, there are two hyperparameters in DropAttention: window size $w$ and drop rate $p$. The window size $w$ is the length of contiguous region to be dropped, while $p$ controls how many units to drop. In standard Dropout \cite{srivastava2014dropout}, the binary mask is sampled from the Bernoulli distribution with the probability of $p$. Since DropAttention will expand every zero entry in the binary mask to be window with size $w$. Therefore, we just require to drop $p/w$ windows.

\subsection{Normalized Rescaling}
To ensure that the sum of attention weights to remain 1 after applying DropAttention, we re-normalize the attention weights after dropout operations. 
While traditional Dropout also has rescaling operation where neuron weights are divided by $1-p$, there is no guarantee that the sum of attention weights after rescaling remains 1. Experiments on classification task (see sec. \ref{sec:tc}) show that DropAttention with normalized rescaling outperforms traditional dropout rescaling. And in practice with normalized rescale, training process can be more steady compared to traditional rescaling.

Figure \ref{fig:dropattention} shows two proposed DropAttention methods. The Pseudocode of DropAttention(e) is described in Algorithm \ref{dropattentionalgorithm}. DropAttention(c) is adopted in the similar way to DropAttention(e).

\begin{algorithm}[t]
    \caption{DropAttention(e)}\label{dropattentionalgorithm}
    \begin{algorithmic}[1]
    \Require Attention weight matrix $\Lambda$, window size $w$, drop rate $p$
        \If{\texttt{Inference}}
            \State{return $\Lambda$}
        \EndIf
        \State{$\gamma=p/w$;}
        \State{Sample mask matrix $M$ randomly, where $M_{ij} \sim Bernoulli(\gamma)$;}
        \State{For each zero position $M_{ij}$, expand the mask with the span length of $w$, $M_{i,j}:M_{i,j+w-1}$, and set all the values in the window to be $0$;}
        \State{Apply mask: $\Lambda =M\odot \Lambda $;}
        \ForAll{row vector of $\Lambda$: $\mathbf{\lambda}_j$}
        \State{Normalized rescale: $\mathbf{\lambda}_j=\mathbf{\lambda}_j/ sum(\mathbf{\lambda}_j)$\; }
        \EndFor
    \end{algorithmic}
\end{algorithm}

\section{Experiment}

We evaluate the effectiveness of DropAttentions on 4 different tasks: Text Classification, Sequence Labeling, Textual Entailment and Machine Translation. We also conduct a set of analytical experiments to validate properties of the networks.

\subsection{Text Classification}\label{sec:tc}

We first evaluate the effectiveness of DropAttention on a couple of classification datasets ranging from small, medium and large scale.
Statistics of datasets are listed in Table \ref{tab: dataset statistics}. All datasets are split into training, development and testing sets.

\begin{table}[!t]\footnotesize
\caption{Classification dataset statistics, \#classes denotes the number of classes, and \#documents represents the number of documents.}
\center
\tabcolsep0.06in
\begin{tabular}{{c}{c}{c}}
\toprule
\textbf{Dataset} &       \textbf{\#classes} &	 	 \textbf{ \#documents } \\
\midrule

CR &2 & 3,993\\
\midrule
 QC &6 & 5,052\\
\midrule
SUBJ &2 & 10,000\\
\midrule
 MR &2 & 10,661\\
 \midrule
 AG's News &4 & 127,600\\
  \midrule
Yelp2013 &5 & 335,018\\

\bottomrule

\end{tabular}
\label{tab: dataset statistics}
\end{table}






\textbf{Yelp13 reviews}: collected from the Yelp Dataset Challenge in 2013, which have 5 levels of ratings from 1 to 5. We use the same Yelp datasets slitted and tokenized in \cite{tang2015document}. \textbf{MR}: Movie reviews with two classes \cite{pang2005seeing}. \textbf{SUBJ}: Subjectivity dataset containing subjective and objective instance. It is also a 2 classes dataset \cite{pang2004sentimental}. \textbf{CR}: Customer reviews of various products with positive and negative sentiments. \textbf{AG's News}: A news topic classification with 4 classes created by \cite{zhang2015character}. \textbf{QC}: The TREC questions dataset involves six different question types \cite{li2002learning}.

\begin{table*}[!t]
\caption{Text classification, in percentage. $p$ represents dropout rate, $w$ represents window size. The column of ``Norm?'' indicates the results of normalized rescaling or traditional rescaling $1-p$. We only represents the best results in the table and their corresponding hyperparameters.}
\center \footnotesize
\tabcolsep0.06in
\begin{tabular}{l*8{c}}
\toprule
  \multirow{1}{*}{ \textbf{Model}}
  &\textbf{Norm?}
& \multirow{1}{*}{\textbf{CR}}
& \multirow{1}{*}{\textbf{SUBJ}}
& \multirow{1}{*}{\textbf{MR}}
& \multirow{1}{*}{\textbf{QC}}
& \multirow{1}{*}{\textbf{AG's News}}
& \multirow{1}{*}{\textbf{Yelp13}}  \\

\midrule
w/o DropAttention    &         &  80.00  & 93.30  & 76.92    &  88.40  & 88.13  & 61.49  \\
 \midrule
& & p=0.4,w=2 & p=0.2,w=3 & p=0.3,w=2 & p=0.3 w=1 & p=0.4 w=1 & p=0.4 w=1 \\
\cmidrule(lr){3-3}   \cmidrule(lr){4-4} \cmidrule(lr){5-5} \cmidrule(lr){6-6} \cmidrule(lr){7-7} \cmidrule(lr){8-8} \cmidrule(lr){9-9}
DropAttention(c)   &  Y & \textbf{82.75}   &  \textbf{94.10} & \textbf{78.80} & \textbf{90.80} & \textbf{88.87} & \textbf{62.34}     \\
& N & 78.25 & 93.10 &  77.30 &  89.60 & 88.49 & 62.27 \\
\midrule
& & p=0.2,w=3 & p=0.3,w=2 & p=0.3,w=2 & p=0.3,w=2 & p=0.2,w=2 & p=0.2 w=1 \\
\cmidrule(lr){3-3}   \cmidrule(lr){4-4} \cmidrule(lr){5-5} \cmidrule(lr){6-6} \cmidrule(lr){7-7} \cmidrule(lr){8-8}

DropAttention(e) &  Y  & 81.25  & 93.50 & 78.51 & 89.60  & 88.66 & 61.79\\
& N & 81.25 & 93.50 & 75.33 &  88.80 & 88.47 & 61.46\\

\bottomrule
\end{tabular}
\label{Classification Results}
\end{table*}


Detail model configurations are given in Appendix. We use accuracy as evaluation metrics. Results of all datasets are listed in Table \ref{Classification Results}. It shows that DropAttentions can significantly improve performance on a wide range of datasets of small, medium and large scale. Besides, note that when comparing normalized rescaling with traditional rescaling under the same DropAttention hyperparameters, Table \ref{Classification Results} shows that normalized rescaling can generally obtain better performance.

For classification tasks, we find that larger dropout rate and smaller window size are preferred for DropAttention(c) while smaller dropout rate and larger window size are preferred for DropAttention(e). And DropAttention(c) can generally obtain higher performances than DropAttention(e) in classification tasks.

\subsection{Sequence Labeling}

We also evaluate the effectiveness of DropAttention on sequence labeling. We conducted experiments by following the same settings as \cite{yang2016multi}.
We use the following benchmark datasets in our experiments: Penn Treebank (PTB) POS tagging, CoNLL 2000 chunking, CoNLL 2003 English NER. The statistics of the datasets are described in Table~\ref{tab:st}.

\begin{table}[!t]\footnotesize
\caption{The sizes of the sequence labeling datasets in our experiments, in terms of the number of tokens.}
\center
\tabcolsep0.06in
\begin{tabular}{l*{6}{l}}
\toprule
\textbf{Dataset} &       \textbf{Task} &	 	 \textbf{Train } &	 \textbf{Dev. } &	 \textbf{Test } \\
\midrule

CoNLL 2000 &\multirow{1}{*}{Chunking}
&  211,727 &  - &  47,377    \\
\midrule
CoNLL 2003 &\multirow{1}{*}{NER}
&  204,567 &  51,578 &  46,666    \\
\midrule
PTB &  \multirow{1}{*}{POS}
&  912,344 &  131,768 &  129,654     \\

\bottomrule
\end{tabular}
\label{tab:st}
\end{table}

We process sentences with Transformer encoder. After encoding, we feed the output vector into a fully-connected layer. Detail model hyperparameters are given in Appendix.

Results are shown in Table \ref{sequence labeling results}. In Table \ref{sequence labeling results}, all best results are under the hyperparameters of $p=0.3,w=3$ except for DropAttention(e) in POS task with $p=0.2,w=2$. It shows that both DropAttention(c) and DropAttention(e) can obtain significant improvements. Our model achieve 0.29 accuracy, 0.40 F1 score, 1.76 F1 score improvements in POS, NER and Chunking respectively. And we find that larger dropout rate and larger window size are generally preferred.






\begin{table}
\parbox{0.51\linewidth}{
    \caption{Sequence labeling results. $p$ means dropout rate, $w$ means window size. NER and Chunking are evaluated by F1 score while POS is evaluated by accuracy. Table shows the best results and their corresponding hyperparameters.}
    \center
    \tabcolsep0.06in
    \resizebox{0.53\columnwidth}{!}{
      \begin{tabular}{l*8{c}}
        \toprule
          \multirow{1}{*}{ \textbf{Transformer}}
        & \multirow{1}{*}{\textbf{POS}}
        & \multirow{1}{*}{\textbf{NER}}
        & \multirow{1}{*}{\textbf{Chunking}}
          \\

        \midrule
        w/o DropAttention    &   95.92      &  87.23     & 89.09  \\
         \midrule
        &  p=0.3 w=3 &p=0.3 w=3 & p=0.3 w=3  \\
        \cmidrule(lr){2-2} \cmidrule(lr){3-3}  \cmidrule(lr){4-4}
        DropAttention(c)   &  \textbf{96.21} & 88.51   &  90.56      \\
        \midrule
        & p=0.2,w=2 & p=0.3,w=3 & p=0.3,w=3 \\
        \cmidrule(lr){2-2} \cmidrule(lr){3-3}   \cmidrule(lr){4-4}
        
        DropAttention(e) & 96.17  & \textbf{88.63}  & \textbf{90.85} \\

        \bottomrule
        \end{tabular}
    }
    \label{sequence labeling results}
}
\hfill
\parbox{0.45\linewidth}{
	\caption{Machine Translation performances of our models under different dropping settings. $p$ stands for drop rate and $w$ represents window size. }
    \center
    \tabcolsep0.06in
    \resizebox{0.43\columnwidth}{!}{
      \begin{tabular}{llclc}
        \toprule
        \multicolumn{1}{c}{\multirow{2}{*}{\textbf{Model}}}    &  \multicolumn{3}{c}{\textbf{HyperParam}} &  \multicolumn{1}{c}{\multirow{2}{*}{\textbf{BLEU}}}  \\
        \cmidrule(lr){2-4}
        &  p  & w  & \\
        \midrule
         w/o DropAttention & 0   &   0  &    & 27.30 \\
         \midrule
        
        \multicolumn{1}{c}{\multirow{6}{*}{DropAttention(c)}} & 0.1        &   1  &    & 27.96 \\
                       &  0.1       &   2  &    &  27.87  \\
                       &  0.1       &   3  &    &  27.98 \\
                       &  0.2       &   1  &    &  27.87  \\
                       &  0.2       &   2  &    &  \textbf{28.04}  \\
                       &  0.2       &   3  &    &  27.95 \\
        
         \midrule
         \multicolumn{1}{c}{\multirow{6}{*}{DropAttention(e)}} & 0.1        &   1  &    & 28.16 \\
                           &  0.1       &   2  &    &  28.03  \\
                           &  0.1       &   3  &    &  28.07  \\
                           &  0.2       &   1  &    &  27.92  \\
                           &  0.2       &   2  &    &  \textbf{28.32}  \\
                           &  0.2       &   3  &    &  27.87  \\

        \bottomrule
        \end{tabular}
    }
    \label{tab:Translation result}
}
\end{table}

\subsection{Textual Entailment}
We use the biggest textual entailment dataset, SNLI \cite{bowman2015large} corpus to evaluate the effectiveness of DropAttention on this task. SNLI is a collection of sentence pairs labeled for entailment, contradiction, and semantic independence. A pair of sentences called premise and hypothesis will be fed to the model, and the model will be asked to tell the relation of two sentences. It is also a classification task, and we measure the performance by accuracy.

We process the hypothesis and premise with the same Transformer encoder, which means that the hypothesis encoder and the premise encoder share the same parameters. We use max pooling to create a simple vector representation from the output of transformer encoder. After processing two sentences respectively, we use the two outputs to construct the final feature vector, which consisting of the concatenation of two sentence vectors, their difference, and their elementwise product \cite{bowman2016fast}. We then feed the final feature vector into a 2-layer ReLU MLP to map the hidden representation into classification result. Detail model hyperparameters are given in Appendix.

Results are listed in Table \ref{tab:SNLI results}. For full results with different hyperparameters please refer to Appendix. Experiments show that DropAttention(c) and DropAttention(e) can significantly improve performances.

\subsection{Machine Translation}\label{sec:mt}
We further demonstrate the effectiveness of DropAttention on translation tasks. We conduct experiments on WMT' 16 En-De dataset which consists of 4.5M sentence pairs. We follow \cite{DBLP:journals/corr/abs-1806-00187} by reusing the preprocessed data, where \cite{DBLP:journals/corr/abs-1806-00187} validates on newstest13 and tests on newstest14, and uses a vocabulary of 32K symbols based on a joint source and target byte pair encoding (BPE; \cite{sennrich2015neural}). We measure case-sensitive tokenized BLEU.
We use the fairseq-py toolkit \footnote{\UrlFont{https://github.com/pytorch/fairseq}} re-implementation of Transformer \cite{vaswani2017attention} model. We follow the configuration of original Transformer base model \cite{vaswani2017attention}. See detail configuration in Appendix. DropAttention with different hyperparameters is applied to attention weights.

Table \ref{tab:Translation result} shows the BLEU score for DropAttention with different hyperparameters.
The results show that DropAttention can generally obtain higher performance compared with baseline without DropAttention. With DropAttention(e) of $p=0.2, w=2$, the model can outperform the baseline by a large margin, reaching a BLEU score of 28.32. For DropAttention(c), the model also reaches the best BLEU score with $p=0.2, w=2$.

There are two insights from this experiment. The first is that a regularization of self-attention works to improve the generalization ability even for the large-scale data.
The second is that the DropAttention is complementary to the standard dropout.

\subsection{Complementarity to stardard Dropout}

We also explore the effect of DropAttention combining with standard Dropout. We conduct experiments on classification tasks and machine translation tasks. We choose AG's News as classification dataset and WMT' 16 En-De as Machine Translation dataset. Same hyperparameters as \ref{sec:tc} and \ref{sec:mt} are used.  Table \ref{complementarity results} shows that when combining DropAttention with Dropout, models can obtain higher performances compared to implementing Dropout or DropAttention alone. It implys that DropAttention is complementary to stardard Dropout.








\begin{table}
\parbox{0.35\linewidth}{
     \caption{SNLI best results and the corresponding hyperparameters.}
    \center
\tabcolsep0.06in
    \resizebox{0.33\columnwidth}{!}{
\begin{tabular}{l*8{c}}
\toprule

  \multirow{1}{*}{ \textbf{Transformer}}
& \multirow{1}{*}{\textbf{SNLI}}

  \\

\midrule
w/o DropAttention    &   83.36        \\
 \midrule
&  p=0.2 w=3    \\
\cmidrule(lr){2-2} 
DropAttention(c)   &  84.38      \\
\midrule
& p=0.5,w=1  \\
\cmidrule(lr){2-2} 

DropAttention(e) &  \textbf{84.48}   \\

\bottomrule
\end{tabular}
    }
    \label{tab:SNLI results}
}
\hfill
\parbox{0.58\linewidth}{
	\caption{Classification and Machine Translation performances. Classification performances are evaluated by accuracy while Machine Translation by BLEU. Baseline is the model without any Dropout techniques.}
	\center
    \tabcolsep0.06in
    \resizebox{0.55\columnwidth}{!}{
          \begin{tabular}{l*6{c}}
    \toprule
      \multirow{1}{*}{ \textbf{Transformer}}
    & \multirow{1}{*}{\textbf{Classification}}
    & \multirow{1}{*}{\textbf{MT}}
      \\

    \midrule
    baseline    &  88.13      &  25.42       \\
     \midrule
     + Standard Dropout   &  88.43 & 27.3        \\
    
     + DropAttention   &  88.50 & 26.3        \\
    
     + Dropout+DropAttention & \textbf{88.70}  & \textbf{28.32}   \\

    \bottomrule
    \end{tabular}
    }
     \label{complementarity results}
}
\end{table}

\section{Analysis}

In this section, we study the impact of DropAttention on the behavior of model quantitatively.
We use three metrics to evaluate the model based on the attention weights: Div, Disagreement and Entropy.

\textbf{Div} Suppose $A$ is the attention weights matrix, where every row i corresponds to the attention weights vector produced by the $i_{th}$ attention head. Div is defined as,
\begin{equation} \label{eu_eqn}
\textrm{Div} =   \left\| \left( A A ^ { T } - I \right) \right\| _ { F } ^ { 2 },
\end{equation}
where $\| \cdot \| _ { F }$ represents the Frobenius norm of a matrix and $I$ stands for identity matrix. It was first introduced by  \cite{lin2017structured} as a penalization term which encourages the diversity of weight vectors across different heads of attention. If Div gets large, it means multi-heads attention weights distributions have large overlap.

\begin{figure*}
\centering
     \subfloat[Entropy]{
\begin{tikzpicture}[scale=1]
  \begin{axis}[
    width = 0.3\linewidth,
    ybar,
    enlargelimits=0.10,
    legend style={at={(0.5,-0.2)},
      anchor=north,legend columns=-1},
    ylabel={Entropy},
    ylabel style={
         yshift=-1ex,
    },
    symbolic x coords={p=0.3 w=3, p=0.3 w=2,p=0.3 w=1,w/o,p=0.2 w=1,p=0.2 w=2, p=0.2 w=3},
    xtick=data,
    x tick label style={rotate=45,anchor=east},
    font=\small,
    ]
    \addplot [draw=blue, fill=blue!15] coordinates { (p=0.3 w=3,1.82786214351654) (p=0.3 w=2,1.82103991508483) (p=0.3 w=1,1.81259489059448)    (w/o,1.81165099143981) (p=0.2 w=1,1.81223118305206)
		(p=0.2 w=2,1.81282675266265) (p=0.2 w=3,1.81787991523742)};
  \end{axis}

  \label{entropy}
\end{tikzpicture}
}
 \subfloat[Disagreement]{
     \begin{tikzpicture}[scale=1]
  \begin{axis}[
  width = 0.3\linewidth,
    ybar,
    enlargelimits=0.10,
    legend style={at={(0.5,-0.2)},
      anchor=north,legend columns=-1},
    ylabel={Disagreement},
    symbolic x coords={p=0.3 w=3, p=0.3 w=2,p=0.3 w=1,w/o,p=0.2 w=1,p=0.2 w=2, p=0.2 w=3},
    xtick=data,
    x tick label style={rotate=45,anchor=east},
    font=\small,
    ]
    \addplot [draw=blue, fill=blue!15] coordinates { (p=0.3 w=3,0.7469131350517273) (p=0.3 w=2,0.7460388541221619) (p=0.3 w=1,0.7456281781196594)    (w/o,0.7439929246902466) (p=0.2 w=1,0.7448163032531738)
		(p=0.2 w=2,0.7450953125953674) (p=0.2 w=3,0.7454040050506592)};
  \end{axis}
  \label{disagreement}
\end{tikzpicture}
}
\subfloat[Div]{
     \begin{tikzpicture}[scale=1]
  \begin{axis}[
  width = 0.3\linewidth,
    ybar,
    enlargelimits=0.10,
    legend style={at={(0.5,-0.2)},
      anchor=north,legend columns=-1},
    ylabel={Div},
    bar width=10pt,
    symbolic x coords={p=0.3 w=3, p=0.3 w=2,p=0.3 w=1,w/o,p=0.2 w=1,p=0.2 w=2, p=0.2 w=3},
    xtick=data,
    x tick label style={rotate=45,anchor=east},
    font=\small,
    ]
    \addplot [draw=blue, fill=blue!15] coordinates { (p=0.3 w=3,0.03334059938788414) (p=0.3 w=2,0.03328307718038559) (p=0.3 w=1,0.03327826038002968)    (w/o,0.03319910541176796) (p=0.2 w=1,0.03320508822798729)
		(p=0.2 w=2,0.033217255026102066) (p=0.2 w=3,0.03324024751782417)};
		
  \end{axis}
  \label{diversity}
\end{tikzpicture}
}
  \caption{The histogram Disagreement, and Div. With the drop rate and window size increasing, both metrics increase accordingly. Note that if the value of Div and Disagreement gets large, it means that the difference of attention weights between heads is small.}
  \label{fig:ana}
\end{figure*}
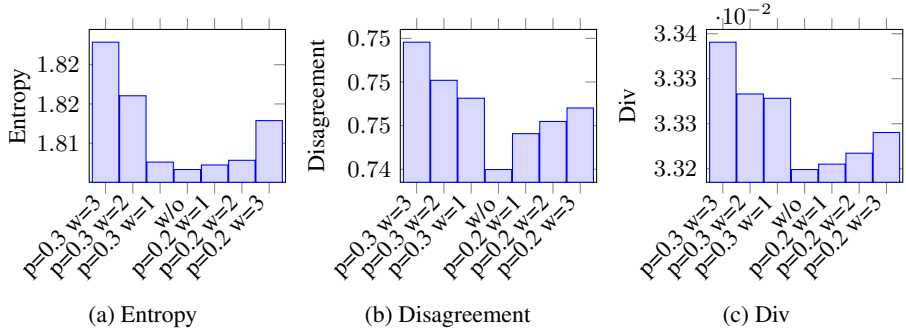

\textbf{Disagreement} We use the same notations above. $A^i$ stands for the $i_{th}$ row of the attention matrix, then the Disagreement is expressed as,
\begin{equation} \label{eu_eqn}
\textrm{Disagreement}= \frac { 1 } { h ^ { 2 } } \sum _ { i = 1 } ^ { h } \sum _ { j = 1 } ^ { h } \frac { A ^ { i } \cdot A^ { j } } { \left\| A ^ { i } \right\| \left\| A ^ { j } \right\| },
\end{equation}
where $h$ denotes the number of heads. It was proposed by \cite{li2018multi}, which also expects to encourage the diversity of the model. The Disagreement is defined as calculating the cosine similarity $\cos ( \cdot )$ between the attention weights vector pair produced by two different heads. The smaller score is, the more diverse different attention heads are.

\textbf{Entropy} is used to evaluate the diversity within one head. $A^i_j$ is the $j_{th}$ element of the attention weights vector produced by $i_{th}$ head. Entropy of attention weights is defined as,
\begin{equation} \label{eu_eqn}
\textrm{E}_i = - \sum_{j} A^i_j \log A^i_j.
\end{equation}
If entropy gets small, it represents that the head focus on a small fraction of words.

\subsection{Effect on Intra-Diversity}

We first observe the impact of DropAttention on intra-diversity, namely attention distribution within one head. Figure \ref{entropy} shows the multi-head entropy of models for classification task. When the drop rate and window size increasing, the entropy increase accordingly. This suggests DropAttention can effectively smoothen the attention distribution, making the model utilize more context. This can subsequently increase robustness of the model.


\subsection{Effect on Inter-Diversity}

We further study the impact of DropAttention on inter-diversity, namely the difference between multi heads. Figure \ref{disagreement} and \ref{diversity} show the Disagreement and Diveristy of multi heads, respectively. It shows that with larger drop rate and window size, Div and Disagreement are larger accordingly. Note that large Diversity and Disagreement means that the difference of attention distribution between heads is small. This is due to the smoother attention distribution within one head. With less sharply different multi-heads, the model does not have to rely on a single head to make predictions, which means that all heads have a smoother contribution to the final predictions. This can increase robustness of the model.

\subsection{Effect on Sparsity}

\begin{figure}[t]
\centering
  \includegraphics[scale=0.45]{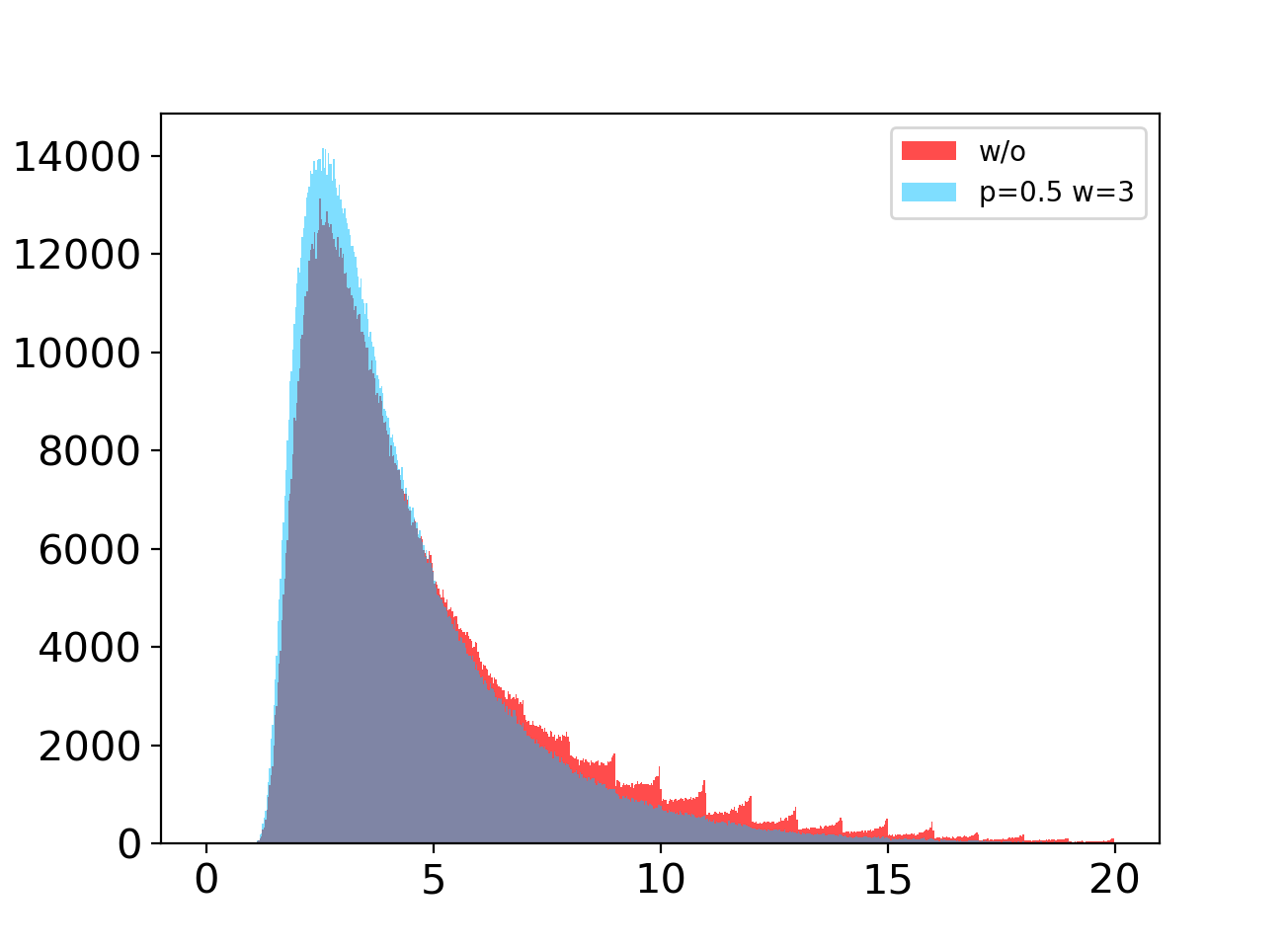}
  \caption{The histogram of largest attention weights distribution. x-axis represents the attention weights value multiplied by the sentence length, y-axis represents the number of corresponding attention weights. Model with DropAttention tends to allocate smaller attention weights compared to model without DropAttention.}
  \label{fig:max-attention}
  \vspace{-2ex}
\end{figure}

Similar to \cite{srivastava2014dropout}, we also observe the effect of DropAttention on sparsity. Since the attention weights are summed up to 1, we only collect the largest attention weights of all heads. To eliminate the effect of sentence length, attention weights are multiplied by the sentence length. Figure \ref{fig:max-attention} shows the distribution of largest attention weights, where model with DropAttention has smaller attention weights compared to model without DropAttention. This phenomenon is consistent with \cite{srivastava2014dropout} where model with dropout tends to allocate smaller activation weights compared to model without dropout.

\section{Conclusion and Discussion}

In this paper, we introduce DropAttention, a variant of Dropout designed for fully-connected self-attention network. Experiments on a wide range of tasks demonstrate that DropAttention is an effective technique for improving generalization and reducing overfitting of self-attention networks. Several analytical statistics give the intuitive impacts of DropAttention, which show that applying DropAttention can help model utilize more context, subsequently increasing robustness.

\bibliography{./nips}
\bibliographystyle{named}

\end{document}